\documentclass[letterpaper]{article} 
\usepackage{aaai25}  
\usepackage{times}  
\usepackage{helvet}  
\usepackage{courier}  
\usepackage[hyphens]{url}  
\usepackage{graphicx} 
\usepackage{natbib}  
\usepackage{caption} 
\usepackage{float}
\usepackage[ruled,vlined]{algorithm2e}

\usepackage{bm}
\usepackage{placeins}
\usepackage{subfigure}
\usepackage{adjustbox}
\usepackage{amsmath}
\usepackage{graphicx}
\usepackage{booktabs}
\usepackage{amssymb}
\usepackage{multirow}
\usepackage{afterpage}
\usepackage[table]{xcolor}
\usepackage{tcolorbox}

\usepackage{newfloat}
\usepackage{listings}
\DeclareCaptionStyle{ruled}{labelfont=normalfont,labelsep=colon,strut=off} 
\floatstyle{ruled}
\newfloat{listing}{tb}{lst}{}
\floatname{listing}{Listing}
\definecolor{myblue}{rgb}{0.36,0.64,0.83}
\usepackage{hyperref}
\hypersetup{
    colorlinks=true,
    urlcolor=red,
    citecolor=myblue,
}

\title{Game4Loc: A UAV Geo-Localization Benchmark from Game Data}
\author{
    Yuxiang Ji$^{1}$\equalcontrib,
    Boyong He$^{1}$\equalcontrib,
    Zhuoyue Tan\textsuperscript{\rm 1},
    Liaoni Wu\textsuperscript{\rm 1,2}\thanks{Corresponding Author.}
}
\affiliations {
    \textsuperscript{\rm 1}Institute of Artifcial Intelligence, Xiamen University\\
    \textsuperscript{\rm 2}School of Aerospace Engineering, Xiamen University\\
    yuxiangji@stu.xmu.edu.cn, wuliaoni@xmu.edu.cn \\
    \vspace{10pt}
    \large{Project Page: \textcolor{magenta}{\url{https://yux1angji.github.io/game4loc}}}
}

\begin{document}

\maketitle


\begin{abstract}
The vision-based geo-localization technology for UAV, serving as a secondary source of GPS information in addition to the global navigation satellite systems (GNSS), can still operate independently in the GPS-denied environment.
Recent deep learning based methods attribute this as the task of image matching and retrieval.
By retrieving drone-view images in geo-tagged satellite image database, approximate localization information can be obtained.
However, due to high costs and privacy concerns, it is usually difficult to obtain large quantities of drone-view images from a continuous area.
Existing drone-view datasets are mostly composed of small-scale aerial photography with a strong assumption that there exists a perfect one-to-one aligned reference image for any query, leaving a significant gap from the practical localization scenario.
In this work, we construct a large-range contiguous area UAV geo-localization dataset named GTA-UAV, featuring multiple flight altitudes, attitudes, scenes, and targets using modern computer games.
Based on this dataset, we introduce a more practical UAV geo-localization task including partial matches of cross-view paired data, and expand the image-level retrieval to the actual localization in terms of distance (meters).
For the construction of drone-view and satellite-view pairs, we adopt a weight-based contrastive learning approach, which allows for effective learning while avoiding additional post-processing matching steps.
Experiments demonstrate the effectiveness of our data and training method for UAV geo-localization, as well as the generalization capabilities to real-world scenarios.
\end{abstract}

%

\section{Introduction}
Vision-based UAV geo-localization, as an independent onboard technology that can work independently of communication systems, enables UAVs to autonomously obtain GPS information even when GNSS communication fails.
This UAV visual localization task could be refered as a special case of cross-view geo-localization~\cite{deuserSample4GeoHardNegative2023, zhengUniversity1652MultiviewMultisource2020, huCVMNetCrossViewMatching2018a} and visual place recognition~\cite{arandjelovicNetVLADCNNArchitecture2016}. 
Recent research formulates this as a cross-view image retrieval problem~\cite{linJointRepresentationLearning2022, daiVisionBasedUAVSelfPositioning2023}.
Given a drone-view image, the goal is to retrieve a matching scene from a database of GPS-tagged satellite-view images to infer the current GPS information of the UAV.
Compared to traditional hand-crafted feature extraction algorithms, deep learning based methods achieve higher accuracy and better generalization performance~\cite{tianCrossViewImageMatching2017, dusmanuD2NetTrainableCNN2019}.
However, such superiority is built upon the training on a large amount of paired images from drone-view and satellite-view.

\begin{figure}[]
    \centering
    \includegraphics[width=\linewidth]{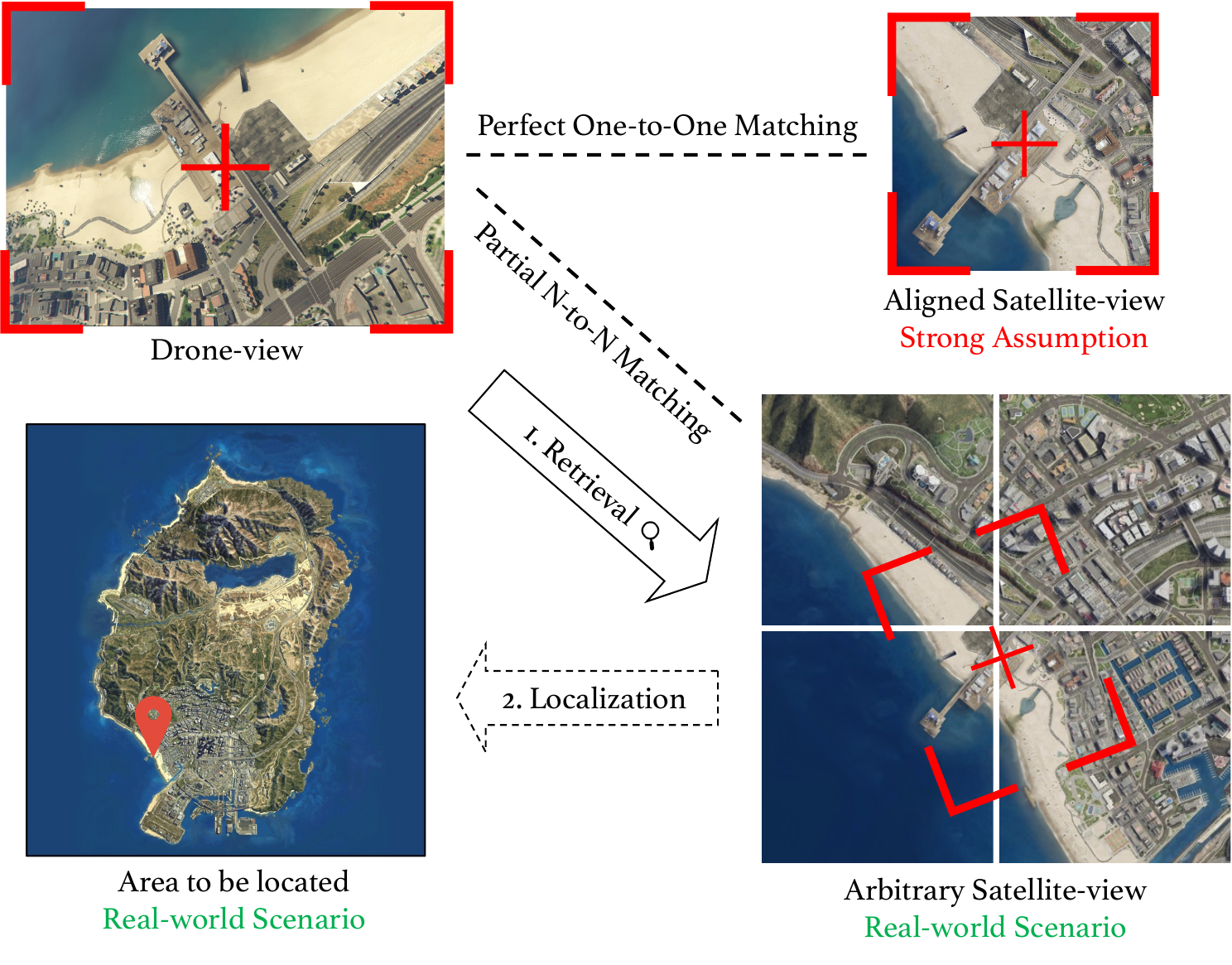} 
    \caption{
        Comparision between perfect matching pair and partial matching pair.
    }
    \label{fig:comparision perfect and partial}
\end{figure}

Existing cross-view datasets are mostly composed of image pairs from different platform views, e.g., ground cameras and satellites~\cite{workmanWideAreaImageGeolocalization2015, zhaiPredictingGroundLevelScene2016, liuLendingOrientationNeural2019}.
The datasets for UAV localization follow this paradigm and expand the view to drones~\cite{zhengUniversity1652MultiviewMultisource2020, xuUAVVisLocLargescaleDataset2024, zhuSUES200MultiheightMultiscene2023, daiVisionBasedUAVSelfPositioning2023}.
Due to high costs and privacy concerns, most of these data are obtained through Google Earth Engine simulation, and the remaining real-world data are very limited in terms of scale, height, angle, etc.
More critically, these datasets simply assume that \textit{each query drone-view image has a perfectly one-to-one ailgned matching satellite-view image as a reference}, which does not apply to practical scenarios because it is impossible to obtain an arbitrary view of drone in advance and align it with a satellite-view reference.
Consequently, such perfect matches are very unlikely to exist in practical scenarios; instead, it is more common to encounter partial matching pairs between drone-view and satellite-view as shown in Fig.~\ref{fig:comparision perfect and partial}.
This leads to models trained on such paradigm datasets struggling to handle practical UAV visual localization tasks.

In fact, some works already noticed the above problems, and attempted to address it from both task desgin and data construction perspectives.
VIGOR~\cite{zhuVIGORCrossViewImage2021} introduces a beyond ont-to-one matching task for ground-satellite matching.
DenseUAV~\cite{daiVisionBasedUAVSelfPositioning2023} and UAV-VisLoc~\cite{xuUAVVisLocLargescaleDataset2024} are two continuous range real-world drone-satellite paired datasets.
Both of them expand the retrieval task to localization; however, the former data construction method still does not align with practical scenarios, and the latter lacks a definition of data pair construction and task design. 
Additionally, these real-world data are limited in terms of scenes, camera angles, and flight altitudes/attitudes, which restricts its generalization performance in diverse scenarios.

In light of the above problems, we propose aligning directly with practical tasks at the data construction level by expanding the original perfect matching to encompass partial matching as Fig.~\ref{fig:comparision perfect and partial}.
Under our setting, the drone-satellite pairs are consturcted following the real-world scenarios, where \textit{drone-view images are retrieved from a gallery of satellite-view images containing partial matches}.
By consturcting such retrieval task, we can recreate the real-world UAV visual geo-localization scenarios from the task design and evaluate the localization performance based on the retrieval results.
Based on this, to replicate various drone flight conditions, we utilize commercial video games to simulate and collect a contiguous large-range of drone-satellite image pairs dataset GTA-UAV from multiple flight alittudes/attitudes, and various flight scenarios.
In total, 33,763 drone-view images are collected from the entire game map, encompassing various scenes such as urban, mountain, desert, forest, field, and coast.

In conjunction with this data consturction method, we introduce a weighted contrastive learning approach weighted-InfoNCE, to utilize the intersection ratio of the partially matched data areas as weight labels for contrastive learning between the paired data.
Exeperiments demonstrate that through this training method, the network can reduce the embedding distance of partially matched samples from different views, making retrieval and localization available.

Our contribution can be summarized as following:
\begin{itemize}
    \item We introduce a new benchmark and dataset for the problem of UAV geo-localization. This dataset, for the first time, expands the perfect matching UAV geo-localization task to include partial matches, allowing for a more realistic task.
    \item We propose a weighted contrastive learning method weighted-InfoNCE to enable the model to learn this partial matching paradigm.
    \item We validate the effectiveness of proposed dataset and method, and demonstrate their potential and generalization capabilities in real-world tasks using a small amount of available real data. 
\end{itemize}

\section{Related Work}
\subsection{Cross-view Geo-Localization Datasets}
Due to the comprehensive coverage of high-altitude reference data such as satellite and aerial imagery, most studies use GPS-tagged satellite imagery as the reference view for cross-view geolocalization.
Among them, many datasets focus on the cross-view matching between ground-level and satellite-view~\cite{linCrossViewImageGeolocalization2013, tianCrossViewImageMatching2017, liuLendingOrientationNeural2019, zhaiPredictingGroundLevelScene2016, zhuVIGORCrossViewImage2021}.
Specifically, VIGOR~\cite{zhuVIGORCrossViewImage2021} doubts the perfect one-to-one matching data pairs and introduces the concept of beyond one-to-one retrieval in ground-satellite matching.
University-1652~\cite{zhengUniversity1652MultiviewMultisource2020} frist introduces the drone-view into the cross-view datasets, where each drone-satellite pair focuses on a target university building.
Although the drone's perspective can serve as a retrieval target, the task still not achieve geolocalization.
In following works, DenseUAV~\cite{daiVisionBasedUAVSelfPositioning2023} and SUES-200~\cite{zhuSUES200MultiheightMultiscene2023} change discrete sampling into continuous sampling and consider different altitudes.
Constrained by flight costs and the limitations of Google Earth simulation, the variety of shooting angles and altitudes reamins very limited.
Most importantly, these datasets construction methods still adhere to the one-to-one perfect matching paradigm and do not align with practical scenarios.
UAV-VisLoc~\cite{xuUAVVisLocLargescaleDataset2024} is a recently released real high-altitude drone dataset where each drone-view image is geotagged, while no clear task desgin has been defined for this data yet.

\subsection{Cross-view Geo-Localization Methods}
One of the first deep learning based geolocalization works by Workman et al.~\cite{workmanWideAreaImageGeolocalization2015} demonstrates the superior accuracy and generalization of CNNs compared to traditional hand-crafted features.
They simply utilize a L2 Loss to minimize the feature distance between cross-views and perform retrieval based on feature distances.
Some works~\cite{linLearningDeepRepresentations2015, voLocalizingOrientingStreet2017, arandjelovicNetVLADCNNArchitecture2016} adopt the idea of contrastive learning, reducing the distance between positive sample pairs.
Yang et al. and Zhu et al.~\cite{yangCrossviewGeolocalizationLayertoLayer2021, zhuTransGeoTransformerAll2022} explore the Transformer architecture in geolocalization to extract additional geometric properties.
Specifically, Chen et al.~\cite{Chen_2024} proposes research on the unaligned case, i.e., the non-centered or shifting targets.
However, their experiments are still conducted on aligned datasets.
Sample4Geo~\cite{deuserSample4GeoHardNegative2023} adopts the recent pre-training approach used in vision-language work CLIP~\cite{radfordLearningTransferableVisual2021}, applying large batch size contrastive learning to cross-view data.
They enhance the learning effect by constructing numerous hard negatives based on InfoNCE~\cite{oordRepresentationLearningContrastive2019}.

\begin{table*}[htbp]
    \centering
    \caption{Comparison between the proposed GTA-UAV dataset and existing datasets for UAV visual geo-localization.}
    \label{tbl:comparision dataset}
    \begin{tabular}{l|c|c|c|c|c}
    \toprule
     & \textbf{University} & \textbf{SUES-200} & \textbf{DenseUAV} & \textbf{UAV-VisLoc} & \textbf{GTA-UAV (proposed)} \\ \midrule
    Drone images & 37,854 & 24,210 & 18,198 & 6,742 & 33,763 \\
    Drone-view GPS locations & Aligned & Aligned & Aligned & - & Arbitrary \\
    Altitude range & $\sim 50m$ & $150m \sim 300m$ & $80m \sim 100m$ & $400m \sim 840m$ & $80m \sim 650m$ \\
    Contiguous area & \texttimes & \texttimes & \checkmark & \checkmark & \checkmark \\
    Evaluation in terms of meters & \texttimes & \texttimes & \texttimes & \checkmark & \checkmark \\
    Multiple attitudes & \checkmark & \texttimes & \texttimes & \texttimes & \checkmark \\
    Multiple scenes & \texttimes & \texttimes & \texttimes & \checkmark & \checkmark \\
    Multiple scales satellite images & \texttimes & \texttimes & \texttimes & - & \checkmark \\
    \bottomrule
    \end{tabular}
\end{table*}

\begin{figure*}[htbp]
    \centering
    \includegraphics[width=\textwidth]{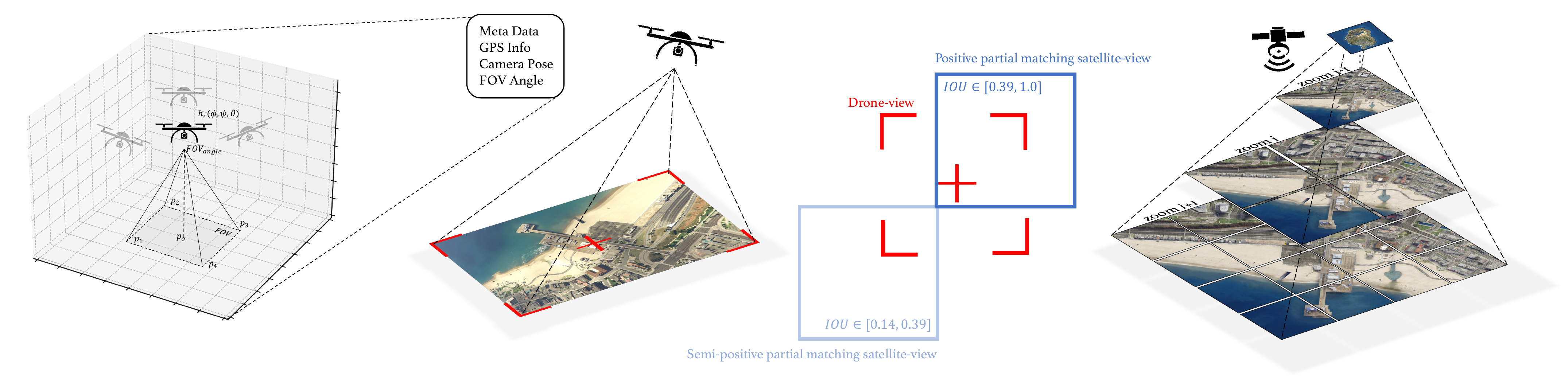} 
    \caption{
        The paired data construction process of GTA-UAV, where \textcolor[rgb]{0.267,0.447,0.769}{Positive} and \textcolor[rgb]{0.710,0.780,0.906}{Semi-positive} satellite-view are paired with \textcolor{red}{Drone-view} by IOU.
    }
    \label{fig:data construction}
\end{figure*}

\section{GTA-UAV Dataset}

\subsection{Problem Statement}
Given a filed of view (FOV) captured by the UAVs, our target is to construct a GPS-tagged reference satellite-view images set from a contiguous area and localize the drone by finding a matching field within it.
Due to the varying flight altitudes and attitudes of UAVs, the FOV can cover multiple scales of the ground area.
To accommodate the varying scales of drone-view, we divide the reference satellite-view of the entire coverage area into multiple hierachical tiles, where the ground resolution between different levels differing by a factor of two.
Unlike the aligned one-to-one retrieval strong assumption of existing datasets in Tab.~\ref{tbl:comparision dataset}, we do not center-align the drone-satellite pairs.
Instead, we use a collect-then-match approach, pairing them by calculating the overlapping of the ground area covered by the two views.
In such arbitrarily sampling way, the relationship between pairs changes from perfectly aligned matching to partial matching.
Refer to the definition of positive samples in VIGOR~\cite{zhuVIGORCrossViewImage2021}, we attribute samples with a ground area intersection over union (IOU) greater than 0.39 as a positive pair, and IOU greater than 0.14 as a semi-positive pair.
The positive pairs are considered as ground truth for retrieval for their highest match, while semi-positive pairs are complementary to the paritial matching learning.
Such paritial matching, in contrast to the strong assumption of perfect matching, can be considered a more challenging retrieval task.
On the basis of coarse retrieval, since each of our view data points is GPS-tagged, we can also evaluate the retrieval results at the distance level.
This provides a foundation for fine localization in further research.
Comparing to the existing datasets for UAV visual geo-localization as Tab.~\ref{tbl:comparision dataset}, our proposed GTA-UAV dataset offers higher flexibility and can cover a wider range of task scenrarios.
We believe that our dataset complements existing UAV visual localization datasets and significantly bridge the gap between current research and practical applications.

\begin{figure*}[!h]
    \centering
    \includegraphics[width=\textwidth]{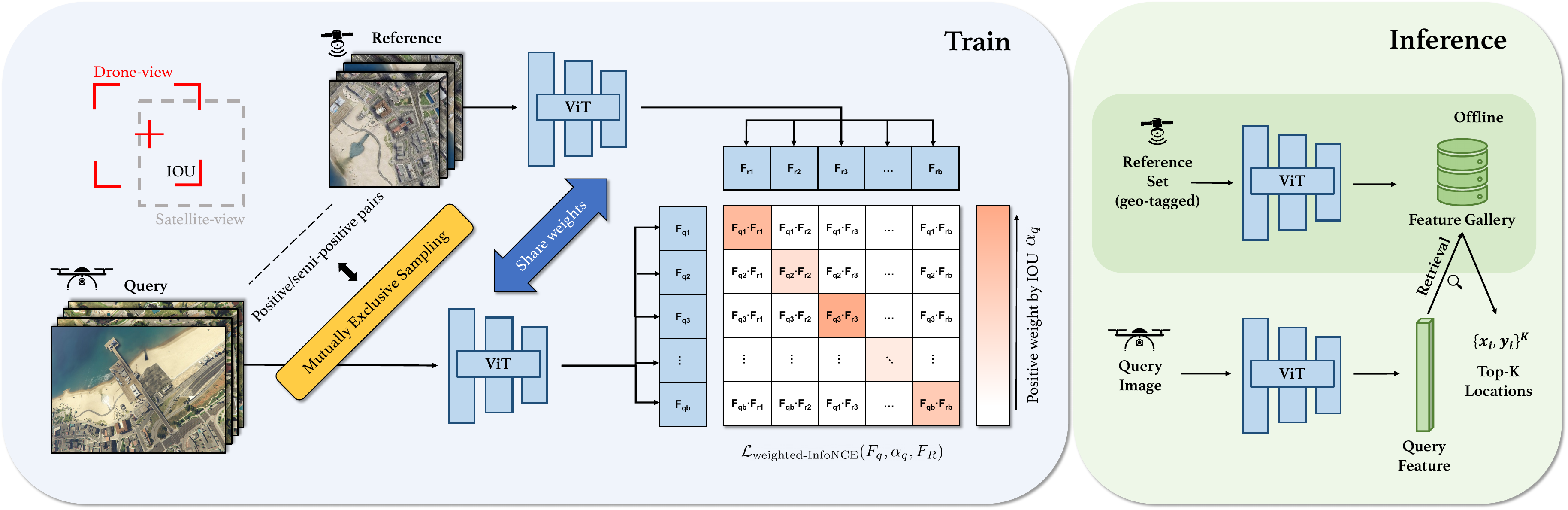} 
    \caption{
        The overview of our training and inference pipeline. 
        (\textbf{left}) We use ViT as feature encoder and weighted-InfoNCE for training positive and semi-positive batched samples from mutually exclusive sampling.
        (\textbf{right}) Then the retrieval could be based on discriminative features to achieve localization.
    }
    \label{fig:pipeline}
\end{figure*}

\subsection{Data Collection and Construction}
In light of the existing works~\cite{richterPlayingDataGround2016,rosSYNTHIADatasetLarge2016,kieferLeveragingSyntheticData2021} on synthetic data, we utilize Grand Thef Auto V (GTAV) as a simulation platform.
We collect 33,763 drone-view images covering distinctive areas in the whole game map, including urban, mountain, desert, forest, field, and coast.
To cover various flight altitudes and attitudes of UAVs, we simulate multiple flight heights ranging from $80m$ to $650m$, and multiple camera angle ranges for roll $\phi \in [-10^\circ, 10^\circ]$, pitch $\theta \in [-100^\circ, -80^\circ]$ and yaw $\psi \in [-180^\circ, 180^\circ]$.
The raw drone-view images are captured in $1920 \times 1440$ with GPS tagged for meter-level evaluation.
Based on the entire game map's area of $81.3km^2$, we utilize a staellite map with a ground resolution of about $0.2m$ and divide it into a total of 8 hierarchical tiles.
Each tile image has a pixel resolution of $256 \times 256$, where the highest zoom level tiles having a ground resolution of about $0.27m$.
We collect totaling 14,640 tiles from zoom levels 4 to 7 as reference satellite-view set, to accommodate possible flight altitudes.
For each drone-view image, we record the GPS information, flight altitude, flight attitude, and camera angle at the time of capture.
By combining the FOV angle setting, we could approximate the ground area covered by the drone-view FOV.
Then by enumerating the nearby satellite tiles from each level for each drone-veiw image, we set those with a ground coverage IOU greater than 0.39 as a positive drone-satellite pair, and the IOU between 0.14 and 0.39 as a semi-positive drone-satellite pair as shown in Fig.~\ref{fig:data construction}.
The detailed construction process and dataset statistics are put in the supplementary.

\subsection{The Evaluation Protocal}
Based on the existing works of geo-localization~\cite{zhuVIGORCrossViewImage2021, daiVisionBasedUAVSelfPositioning2023, zhengUniversity1652MultiviewMultisource2020}, we utilize two retrieval-based metrics (Recall@K, AP) and one localization-related metric (SDM@K~\cite{daiVisionBasedUAVSelfPositioning2023}) for evaluation.
In addition, we include distance error between the retrieval results and the query location as an evaluation method.
Based on this, we introduce two application scenarios as the same in VIGOR~\cite{zhuVIGORCrossViewImage2021}: same area and cross area.
The same area represents the scenario where both the training and the testing data pairs are sampled from the same area, reflecting applications where the flight area data is available.
The cross area represents the case that the training and testing data are seperated.
Under this setting, we divide half of the game map into training data and evaluate on the other half, and these areas differ on the scenes.

\section{Geo-localization via Cross-view Matching}

\subsection{Baseline Framework}
Large-scale UAV geo-localization necessitates a trade-off between accuracy and performance.
Practical application scenarios demand that the pipeline avoids complex pre-processing and post-processing steps.
We avoid introducing additional matching modules in the retrieval-based paradigm, allowing the reference statellite-view set to be processed offline and retrieval to be performed through simple distance similarity measures.
Recent works typically use a Siamese Network to encode cross-view images and train a model for generating cross-view descriptors using Triplet loss or some variant of metric learning~\cite{deuserSample4GeoHardNegative2023, voLocalizingOrientingStreet2017, huCVMNetCrossViewMatching2018a, liJointlyOptimizedGlobalLocal2023, zhuTransGeoTransformerAll2022}.
To simplify the entire pipeline and align with the model structure of standard visual tasks for simply comparing different data pre-training effects, we directly utilize a pair of weight-sharing original Vision-Transformer (ViT) models~\cite{dosovitskiyImageWorth16x162021} with default Multi-Layer Perceptron (MLP) head as the descriptor model, without introducing any additional fusion modules.
We follow the training approach using Symmetric InfoNCE from Sample4Geo~\cite{deuserSample4GeoHardNegative2023} as the baseline, leveraging all available negatives in batch learning.

\subsection{Weighted Positive Training}
\label{sec:weight positive training}
Directly utilizing the original Triplet loss or symmetric InfoNCE loss allows the constructed paired data to be treated as positive samples and non-paired data as negative samples for contrastive learning.
This approach works well in one-to-one perfect matching pairs.
However, in our arbitrary partial matching paired data, treating all degrees of partial matching as equal-weight positive samples could introduce significant bias, affecting the learning result and training stability.
Based on our data consturction method, we utilize the IOU of ground area covered by cross-view pairs $\text{IOU}_{qr^+}$ as additional supervision information for contrastive learning as:
\begin{equation}
    \begin{split}
        \mathcal{L}_\text{weighted-InfoNCE}(F_q,\alpha_q,F_R) = \\
        &\hspace{-2em} -\alpha_q \log \frac{\exp(F_q \cdot F_{r^+}/\tau)}{\sum_{i}^R \exp(F_q \cdot F_{r_i} /\tau)} \\
        &\hspace{-7em} -(1 - \alpha_q) \frac{1}{|R|}\sum_{i}^R \log \frac{\exp(F_q \cdot F_{r_i^{+,-}}/\tau)}{\sum_{j}^R \exp(F_q \cdot F_{r_j}/\tau)} \\
        &\hspace{-13em}= \alpha_q \mathcal{L}_\text{InfoNCE}(F_q,F_R) + (1 - \alpha_q)\mathcal{L}_\text{uniform-InfoNCE}(F_q,F_R),
    \end{split}
\label{eq:weighted infonce}
\end{equation}
where $F_q$ represents an encoded query image from one-view, $F_R$ represents the encoded reference images from another view in the same batch, and $r^+$ represents positive/semi-positive reference pair.
The $\tau$ denotes a learnable parameter~\cite{radfordLearningTransferableVisual2021}.
The weight coefficients $\alpha_q$ are calculated by parametric Sigmoid as Eq.~\ref{eq:weight alpha}:
\begin{equation}
    \alpha_{q} = \sigma(k, \text{IOU}_{qr^+}) = \frac{1}{1 + \exp(-k \times \text{IOU}_{qr^+})},
\label{eq:weight alpha}
\end{equation}
where $k$ is a hyper-parameter and higher value represents greater curvature change.
When $k$ approaches infinity, the loss function degenerates into the standard InfoNCE.
In a single batch of size $N$, there are $N$ positive/semi-positive paired samples with corresponding positive weights from two views, and the loss function as Eq.~\ref{eq:weighted infonce} would be calculated twice symmetrically in two directions (drone to satellite, satellite to drone).
The dot-production is utilized as the similarity measurement, where positive/semi-positive samples are pushed towards higher values.
Building on the original InfoNCE, we incorporate weights for positive/semi-positive sample pairs into the loss function, introducing a degree of flexibility.
This allows the model to adapt the similarity loss based on the extend of partial matching.

\subsection{Mutually Exclusive Sampling}
In the training process based on symmetric InfoNCE introduced in above sections, to establish the negative relationship between sample pairs, we need to sample $N$ pairs of mutually independent positive sample pairs within each batch.
Since there is no guaranteed one-to-one relationship between drone and satellite views in our arbitrary partial matching data construction process, each view image could have neighboring relationships with multiple cross-view images.
In this situation, to adapt to the training pipeline, we utilize a mutually exclusive sampling method as Alg.~\ref{alg:multually exclusive sampling}.
By considering each view image as a node in graph theory and the matching relation as an undirected edge, for each batch, we remove the sampled nodes and all their adajacent nodes.
We then continue sampling from the remaining graph set to avoid having related cross-view data within the same batch.

\section{Experiments}

\begin{table*}[!h]
    \centering
    \caption{
        Performance on GTA-UAV comparing to different training methods.
        MES means Mutual Exclusive Sampling.
    }
    \label{tab:results diff methods}
    \resizebox{\textwidth}{!}{
    \begin{tabular}{l|ccccc|ccccc}
        \toprule
        \multirow{2}{*}{\centering Methods} & \multicolumn{5}{c}{Cross-Area} & \multicolumn{5}{|c}{Same-Area} \\
        \cmidrule(lr){2-6} \cmidrule(lr){7-11}
        & R@1$\uparrow$ & R@5$\uparrow$ & AP$\uparrow$ & SDM@3$\uparrow$ & Dis@1$\downarrow$ & R@1$\uparrow$ & R@5$\uparrow$ & AP$\uparrow$ & SDM@3$\uparrow$ & Dis@1$\downarrow$\\
        \midrule
        \textbf{Positive-only} &&&&&&&&&&\\
        \textcolor{gray}{Triplet Loss ($\mathcal{L}_\text{triplet}$)} & \textcolor{gray}{43.41\%} & \textcolor{gray}{66.70\%} & \textcolor{gray}{53.56\%} & \textcolor{gray}{61.26\%} & \textcolor{gray}{756.95m} & \textcolor{gray}{68.22\%} & \textcolor{gray}{87.99\%} & \textcolor{gray}{76.73\%} & \textcolor{gray}{79.17\%} & \textcolor{gray}{438.38m} \\
        \textcolor{gray}{InfoNCE Loss ($\mathcal{L}_\text{InfoNCE}$)} & \textcolor{gray}{49.57\%} & \textcolor{gray}{72.84\%} & \textcolor{gray}{59.68\%} & \textcolor{gray}{65.53\%} & \textcolor{gray}{612.22m} & \textcolor{gray}{72.99\%} & \textcolor{gray}{90.64\%} & \textcolor{gray}{80.76\%} & \textcolor{gray}{80.40\%} & \textcolor{gray}{363.67m} \\
        \textcolor{gray}{InfoNCE Loss ($\mathcal{L}_\text{InfoNCE}$, w/. MES)} & \textcolor{gray}{52.64\%} & \textcolor{gray}{74.63\%} & \textcolor{gray}{62.40\%} & \textcolor{gray}{67.64\%} & \textcolor{gray}{552.90m} & \textcolor{gray}{72.34\%} & \textcolor{gray}{91.42\%} & \textcolor{gray}{80.86\%} & \textcolor{gray}{81.57\%} & \textcolor{gray}{369.59m} \\
        \textcolor{gray}{Ours ($\mathcal{L}_\text{weighted-InfoNCE}$, w/. MES)} & \textcolor{gray}{\textbf{57.52\%}} & \textcolor{gray}{80.10\%} & \textcolor{gray}{\textbf{67.24\%}} & \textcolor{gray}{72.33\%} & \textcolor{gray}{444.13m} & \textcolor{gray}{75.97\%} & \textcolor{gray}{94.53\%} & \textcolor{gray}{83.35\%} & \textcolor{gray}{82.80\%} & \textcolor{gray}{325.61m} \\
        \midrule
        \textbf{Positive + Semi-positive} &&&&&&&&&&\\
        Triplet Loss ($\mathcal{L}_\text{triplet}$) & 24.78\% & 46.99\% & 35.13\% & 58.79\% & 879.06m & 46.55\% & 85.07\% & 62.95\% & 83.63\% & 252.88m \\
        InfoNCE Loss ($\mathcal{L}_\text{InfoNCE}$) & 35.83\% & 63.79\% & 48.08\% & 68.15\% & 576.41m & 52.67\% & 90.75\% & 67.74\% & 85.35\% & 204.08m \\
        InfoNCE Loss ($\mathcal{L}_\text{InfoNCE}$, w/. MES) & 45.97\% & 71.43\% & 57.19\% & 71.48\% & 460.08m & 65.89\% & 93.09\% & 77.84\% & 86.52\% & 196.59m \\
        Ours ($\mathcal{L}_\text{weighted-InfoNCE}$, w/. MES) & 55.91\% & \textbf{81.07\%} & 66.56\% & \textbf{76.35\%} & \textbf{342.05m} & \textbf{84.95\%} & \textbf{97.59\%} & \textbf{90.15\%} & \textbf{88.03\%} & \textbf{149.07m} \\
        \bottomrule
    \end{tabular}
    }
\end{table*}

\begin{figure*}[!h]
    \centering
    \subfigure{
        \includegraphics[width=0.4\textwidth]{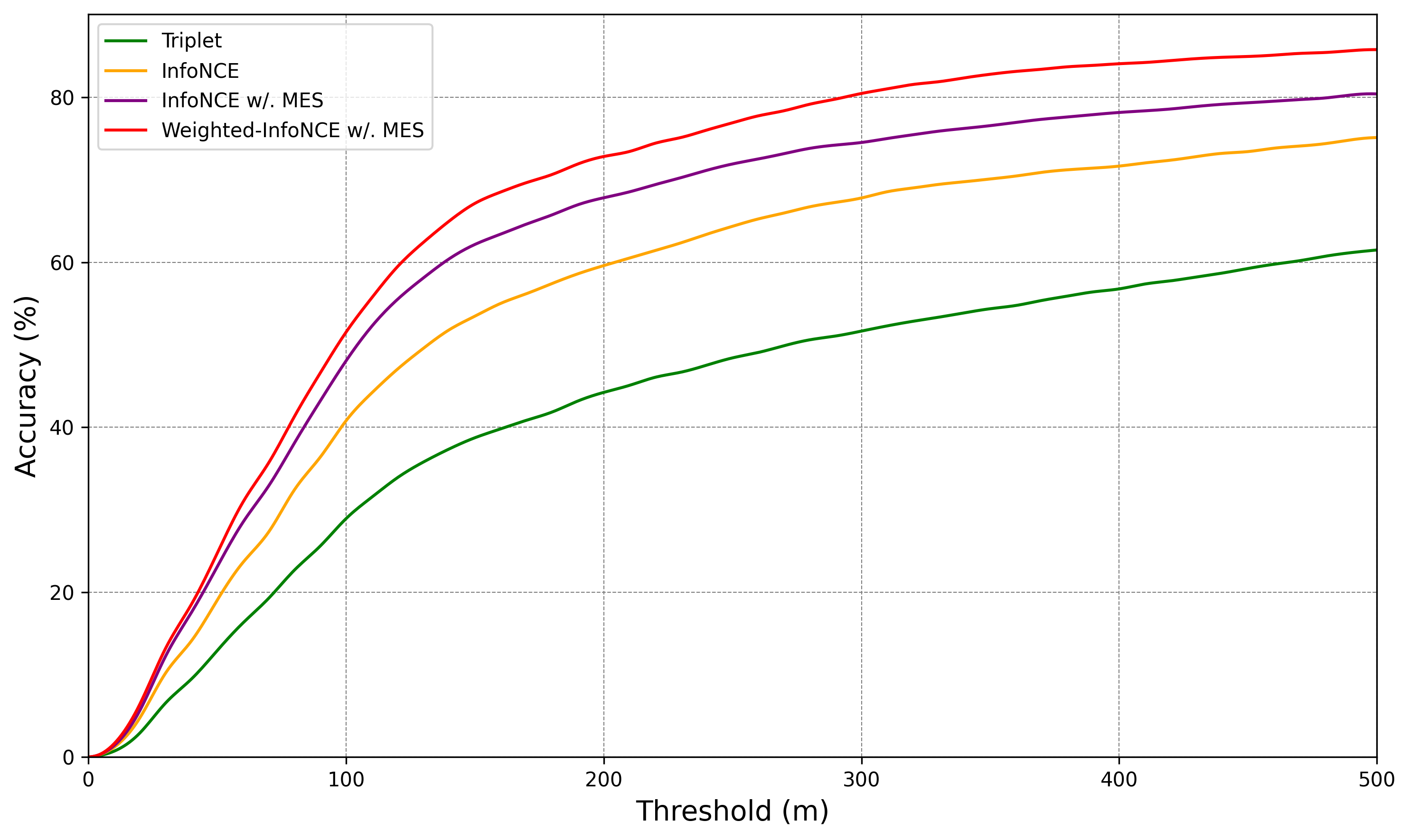}
    }
    \subfigure{
        \includegraphics[width=0.4\textwidth]{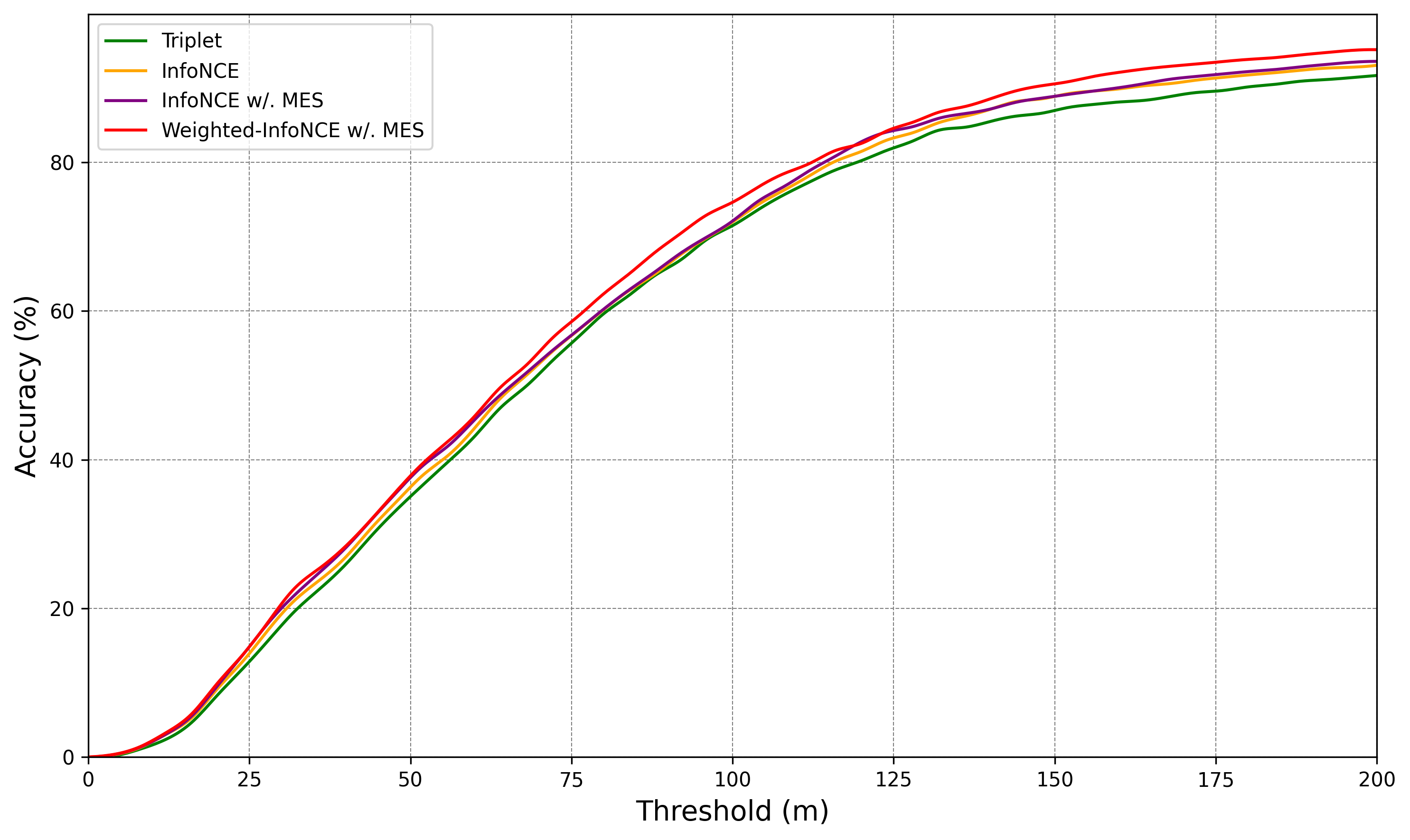}
    }
    \caption{Meter-level localization accuracy of different methods on (\textbf{left}) cross-area and (\textbf{right}) same-area.}
    \label{fig:meter-level localization accuracy}
\end{figure*}
\begin{table*}[!h]
    \centering
    \caption{Performance on GTA-UAV comparing to different pre-training datasets.}
    \label{tab:results diff pretrain}
    \resizebox{\textwidth}{!}{
    \begin{tabular}{l|ccccc|ccccc}
        \toprule
        \multirow{2}{*}{\centering Pre-train datasets} & \multicolumn{5}{c}{Cross-Area} & \multicolumn{5}{|c}{Same-Area} \\
        \cmidrule(lr){2-6} \cmidrule(lr){7-11}
        & R@1$\uparrow$ & R@5$\uparrow$ & AP$\uparrow$ & SDM@3$\uparrow$ & Dis@1$\downarrow$ & R@1$\uparrow$ & R@5$\uparrow$ & AP$\uparrow$ & SDM@3$\uparrow$ & Dis@1$\downarrow$\\
        \midrule
        ImageNet~\cite{deng2009imagenet} & 9.74\% & 21.73\% & 15.74\% & 33.58\% & 1841.30m & 10.65\% & 23.90\% & 17.15\% & 36.82\% & 1470.50m \\
        \midrule
        \textbf{Perfect Matching} &&&&&&&&&&\\
        University-1652~\cite{zhengUniversity1652MultiviewMultisource2020} & 32.16\% & 54.19\% & 41.79\% & 54.07\% & 991.64m & 30.90\% & 51.88\% & 40.08\% & 51.62\% & 1166.06m \\
        SUES-200~\cite{zhuSUES200MultiheightMultiscene2023} & 35.29\% & 56.85\% & 44.85\% & 55.32\% & 920.62m & 32.24\% & 52.63\% & 41.38\% & 52.58\% & 1138.93m \\
        DenseUAV~\cite{daiVisionBasedUAVSelfPositioning2023} & 12.89\% & 23.03\% & 17.85\% & 32.33\% & 1848.47m & 12.14\% & 22.06\% & 17.11\% & 30.25\% & 2115.03m \\
        \midrule
        \textbf{Partial Matching} &&&&&&&&&&\\
        GTA-UAV & \textbf{55.91\%} & \textbf{81.07\%} & \textbf{66.56\%} & \textbf{76.35\%} & \textbf{342.05m} & \textbf{84.95\%} & \textbf{97.59\%} & \textbf{90.15\%} & \textbf{88.03\%} & \textbf{149.07m} \\
        \bottomrule
    \end{tabular}
    }
\end{table*}

\subsection{Implementation Details}
In our exeperiments the ViT-Base~\cite{dosovitskiyImageWorth16x162021} with patch-size $16\times16$ and $64M$ parameters is adopted as the image encoding architecture.
Both drone-view images and satellite-view images are resized to $384 \times 384$ before feeding into the network.
The hyper-parameter $k$ of weighted-InfoNCE is set to 5 as default, and the learnable temperature parameter $\tau$ is initialized to $1$.
Following Sample4Geo~\cite{deuserSample4GeoHardNegative2023}, we employ Adam optimizer~\cite{kingmaAdamMethodStochastic2017} with a initial learning rate of 0.0001 and a cosine learning rate scheduler to train each experiment for 20 epochs in batch size of 64.
The flipping, rotation, and grid dropout are included as data augmentation for training.
Both positive and semi-positive pairs are used for training by default if not specifically noted, and we conduct experiments on this in the subsequent subsections.
The further details are put in the supplementary.

\begin{algorithm}
    \caption{Mutually Exclusive Sampling process}
    \label{alg:multually exclusive sampling}
    \KwData{partial paired data $E = \{(q_1, r_1), (q_2, r_2), \ldots, (q_N, r_N)\}$, batch size $b$}
    \KwResult{exclusive batched data $D = \{\{q,r\}^b, ...\} $}
    Initialize $D=\emptyset, D_\text{batch}=\emptyset, G_\text{stack}=\emptyset, G_\text{remain}=E$\;
    \For{$i \leftarrow 1$ \KwTo $N/b$}{
        \For{$e \in G_\text{remain}$}{
            $q_i, r_i \leftarrow e$\;
            $D_\text{batch} \leftarrow D_\text{batch} \cup (q_i, r_i)$\;
            \For{$q_i, r_j \leftarrow E[q_i]$}{
                $G_\text{remain} \leftarrow G_\text{remain} \setminus (q_i, r_j)$\;
                $G_\text{stack} \leftarrow G_\text{stack} \cup (q_i, r_j)$\;
            }
            \For{$q_j, r_i \leftarrow E[r_i]$}{
                $G_\text{remain} \leftarrow G_\text{remain} \setminus (q_j, r_i)$\;
                $G_\text{stack} \leftarrow G_\text{stack} \cup (q_j, r_i)$\;
            }
            \If{$\text{len}(D_\text{batch}) = b$}{
                $D \leftarrow D \cup D_\text{batch}$\;
                $D_\text{batch} \leftarrow \emptyset$\;
                $G_\text{remain} \leftarrow G_\text{remain} \cup G_\text{stack}$\;
                $G_\text{stack} \leftarrow \emptyset$\;
            }
        }
    }
    \Return $D$\;
\end{algorithm}

\subsection{Evaluation Metrics}
For each drone-view query, the top-K images with the highest cosine similarity in the feature embedding space from the satellite-view database would be considered as the retrieval results.
Following the previous works~\cite{deuserSample4GeoHardNegative2023, zhengUniversity1652MultiviewMultisource2020,zhuVIGORCrossViewImage2021}, we first evaluate the retrieval task by Recall@K (R@K) and average precision (AP).
We also include Spatial Distance Metric SDM@K~\cite{daiVisionBasedUAVSelfPositioning2023} as the combined metric for retrieval and localization to further evaluate the positioning performance, where the calculation method is provided in the supplementary.
Considering the average number of references a query may match, we use SDM@3 here.
More intuitively, we provide the distance between the location of the top-$1$ retrieval result and the location of the drone-view query (Dis@1) as an evaluation metric.

\subsection{GTA-UAV Dataset Benchmark}
\label{sec:gta-uav benchmark}
For our GTA-UAV dataset, we compare the proposed method with previous SOTA training methods under both cross-area and same-area settings using positive $+$ semi-positive and positive-only as training data respectively.
As results in Tab.~\ref{tab:results diff methods}, in the proposed paritial matching settings, our proposed weighted-InfoNCE achieves the best results across all metrics.
Specifically, comparing to the previous SOTA method~\cite{deuserSample4GeoHardNegative2023} using InfoNCE, our method improves the R@1 for 20.08\%, and Dis@1 for 234.36m in the cross-area setting trained on positive $+$ semi-positive data.
The results trained on positive $+$ semi-positive data have less retrieval accuracy comparing to the results only trained on positive data.
This is because that the retrieval evaluation considers only the positive references as the correct result, which is precisely the training target of the positive data.
However, for the localization task, the results trained on both positive and semi-positive data achieve better results in the SDM@3 and Dis@1 metrics.
This is because the semi-positive data enable the model to learn a more comprehensive understanding of partial matching relationships.
The further analysis of proposed weighted-InfoNCE are put in the supplementary.

In the above sections, we discuss about the significance of the unaligned partial N-to-N matching paradigm for real-world scenarios.
Here we categorize the existing UAV geo-localization datasets as perfect matching data, and compare the performance of models pre-trained on these perfect matching datasets with their performance on our proposed partial matching GTA-UAV dataset.
The results in Tab.~\ref{tab:results diff pretrain} demonstrate a significant gap between these two tasks, and highlight the substantial importance of our proposed GTA-UAV data for more practical partial matching tasks.

\begin{table*}[!h]
    \centering
    \caption{Transfer performance on UAV-VisLoc with same-area setting comparing different pre-training datasets.}
    \label{tab:results transfer}
    \begin{tabular}{cccccccc}
        \toprule
        \multirow{2}{*}{\centering Exp. Setup} & \multirow{2}{*}{Pre-training datasets} & \multicolumn{5}{c}{Same-Area} \\
        \cmidrule(lr){3-7}
        & & R@1$\uparrow$ & R@5$\uparrow$ & AP$\uparrow$ & SDM@3$\uparrow$ & Dis@1$\downarrow$ \\
        \midrule
        zero-shot & ImageNet~\cite{deng2009imagenet} & 8.35\% & 16.47\% & 13.16\% & 26.53\% & 2615.08m\\
        zero-shot & University-1652~\cite{zhengUniversity1652MultiviewMultisource2020} & 9.61\% & 19.70\% & 14.73\% & 31.67\% & 2285.08m\\
        zero-shot & SUES-200~\cite{zhuSUES200MultiheightMultiscene2023} & 16.71\% & 27.84\% & 22.93\% & 34.07\% & 1959.02m\\
        zero-shot & DenseUAV~\cite{daiVisionBasedUAVSelfPositioning2023} & 18.79\% & 27.09\% & 23.65\% & 32.95\% & 2051.58m\\
        zero-shot & GTA-UAV & \textbf{24.94\%} & \textbf{42.59\%} & \textbf{33.15\%} & \textbf{41.40\%} & \textbf{1689.24m}\\
        \midrule
        fine-tune & ImageNet~\cite{deng2009imagenet} & 74.41\% & 92.36\% & 83.29\% & 80.94\% & 166.63m\\
        fine-tune & University-1652~\cite{zhengUniversity1652MultiviewMultisource2020} & 73.91\% & 93.10\% & 82.05\% & 82.01\% & 170.23m\\
        fine-tune & SUES-200~\cite{zhuSUES200MultiheightMultiscene2023} & 74.44\% & 92.61\% & 81.95\% & 82.10\% & 150.22m\\
        fine-tune & DenseUAV~\cite{daiVisionBasedUAVSelfPositioning2023} & 77.09\% & 92.61\% & 83.82\% & 82.05\% & 139.34m\\
        fine-tune & GTA-UAV & \textbf{80.20\%} & \textbf{96.53\%} & \textbf{87.83\%} & \textbf{85.46\%} & \textbf{122.87m}\\
        \bottomrule
    \end{tabular}
\end{table*}

\begin{table}[!h]
    \centering
    \caption{Performance on GTA-UAV of different models.}
    \label{tab:results diff model}
    \resizebox{\linewidth}{!}{
    \begin{tabular}{lccccccc}
        \toprule
        Model & R@1$\uparrow$ & AP$\uparrow$ & SDM@3$\uparrow$ & Dis@1$\downarrow$ \\
        \midrule
        \textbf{Cross-Area} \\
        ResNet-101 & 13.74\% & 23.06\% & 48.06\% & 1126.52m \\
        ConvNeXt-Base & 55.36\% & 66.14\% & 74.91\% & 386.35m \\
        Swinv2-B & 53.70\% & 65.13\% & \textbf{77.07\%} & 343.30m \\
        ViT-Base/16 & \textbf{55.91\%} & \textbf{66.56\%} & 76.35\% & \textbf{342.05m} \\
        \midrule
        \textbf{Same-Area} \\
        ResNet-101 & 58.10\% & 69.98\% & 82.64\% & 371.78m \\
        ConvNeXt-Base & 83.94\% & 89.54\% & 87.98\% & 160.49m \\
        Swinv2-B & 81.73\% & 88.32\% & 87.35\% & 196.06m \\
        ViT-Base/16 & \textbf{84.95\%} & \textbf{90.15\%} & \textbf{88.03\%} & \textbf{149.07m} \\
        \bottomrule
    \end{tabular}
    }
\end{table}

\subsection{GTA-UAV Transfer Capability}
To further demonstrate the significance of the proposed GTA-UAV dataset for real-world application scenarios, we evaluate the transferability of its pre-trained model to real data with limited number and scenarios.
We select a recently released drone-view dataset, UAV-VisLoc~\cite{xuUAVVisLocLargescaleDataset2024}, which lacks data pairing and task design, as real data.
It includes 6,742 high-altitude, downward-facing images from UAVs, covering several continuous area, and each image is GPS-tagged.
These settings are included in the GTA-UAV dataset, making it a suitable target subset to evaluate the transferability of our dataset.
By using the same data construction method as GTA-UAV, we pair the hierarchical satellite-view images from seven regions and apply identical training and evaluation settings.
The detailed experiment setup and implementations are put in the supplementary.
As shown in Tab.~\ref{tab:results transfer}, comparing to ImageNet, University, SUES-200, and DenseUAV, the model pre-trained on GTA-UAV shows the best zero-shot performance on real UAV geo-localization dataset with cross-area setting. 
Specifically, the R@1 is 6.15\% higher than the second-best result, and the AP is 9.5\% higher.
Similarly, after fine-tuning on UAV-VisLoc, the model pre-trained on GTA-UAV still maintains the highest performance, where the distance error of top-1 retrieval Dis@1 is reduced by 16.47m.

\section{Ablation Study}

\subsection{Architecture Evaluation}
In existing cross-view geo-localization~\cite{deuserSample4GeoHardNegative2023,huCVMNetCrossViewMatching2018a,tokerComingEarthSatellitetoStreet2021,zhuTransGeoTransformerAll2022} research, CNNs and Transformers are widely explored for learning useful representations.
Some studies make adaptive modifications to achieve better learning capabilities~\cite{zhuTransGeoTransformerAll2022,huCVMNetCrossViewMatching2018a,zhuSimpleEffectiveGeneral2023}.
Unlike previous tasks, in the GTA-UAV cross-area task and its corresponding real-world scenarios, the generalization to unseen data in unkown scenes needs to be emphasized.
Based on studies of model generalization~\cite{hoyerDomainAdaptiveGeneralizable2023,jiDiffusionFeaturesBridge2024} and previous SOTA geo-localization methods~\cite{deuserSample4GeoHardNegative2023,zhuTransGeoTransformerAll2022}, we compare several standard architectures in Tab.~\ref{tab:results diff model}.
The results show that the ViT has the best performance under the same order of magnitude parameters.
The practical commonly used architecture ResNet exhibits poor generalization ability, which may be attributed to its relatively weak representational and generalization capacities when dealing with significant variations in displacement, angles, and scenes.
We also conduct experiments on the scale of model parameters in the supplementary.

\subsection{Hyper-parameter Evaluation}
We evaluate different hyper-parameter value $k$ of proposed weighted InfoNCE in Tab.~\ref{tab:results diff k}.
There is a trade-off between treating partial matches as fully positive and maintaining flexibility (controlled by $k$), while all these results outperform when $k \to \infty$ (i.e., the standard InfoNCE).
In addition, considering that the form of weighted-InfoNCE can be regarded as a weight-based label smoothing variant of InfoNCE, we also compare the results of InfoNCE with different fixed smooth value $\epsilon$ in the supplementary.

\begin{table}[]
    \centering
    \caption{Performance on GTA-UAV comparing different hyper-parameters.}
    \label{tab:results diff k}
    \resizebox{\linewidth}{!}{
    \begin{tabular}{lccccc}
        \toprule
        Exp. Setup & R@1$\uparrow$ & R@5$\uparrow$ & AP$\uparrow$ & SDM@3$\uparrow$ & Dis@1$\downarrow$ \\
        \midrule
        \textbf{Cross-Area} \\
        $k=1$ & 52.96\% & 79.62\% & 64.82\% & 74.94\% & 386.56m \\
        $\bm{k=5}$ & \textbf{55.91\%} & \textbf{81.07\%} & \textbf{66.56\%} & \textbf{76.35\%} & \textbf{342.05m} \\
        $k=20$ & 51.50\% & 77.17\% & 62.55\% & 74.16\% & 411.12m \\
        $k \to \infty$ & 45.97\% & 71.43\% & 57.19\% & 71.48\% & 460.08m \\
        \midrule
        \textbf{Same-Area} \\
        $k=1$ & 79.53\% & 97.35\% & 88.79\% & 87.91\% & 173.66m \\
        $\bm{k=5}$ & \textbf{84.95\%} & \textbf{98.53\%} & \textbf{90.15\%} & \textbf{88.03\%} & \textbf{149.07m} \\
        $k=20$ & 77.31\% & 96.80\% & 83.91\% & 87.03\% & 189.73m \\
        $k \to \infty$ & 65.89\% & 93.09\% & 77.84\% & 86.52\% & 196.59m \\
        \bottomrule
    \end{tabular}
    }
\end{table}

\vspace{-0.2cm}
\section{Conclusion}
We propose a new benchmark GTA-UAV for UAV geo-localization with partial matching pairs, which is a more practical setting.
A weighted InfoNCE loss is introduced to leverage the supervision of matching extends.
Extensive experiments validate the effectiveness of our data and method for UAV geo-localization and demonstrate the potential in real-world scenarios.
This work provides a paradigm aligned with real-world tasks for future research.


\bibliography{aaai25}

\end{document}